\pdfoutput=1

\documentclass[11pt,a4paper]{article}
\usepackage[svgnames]{xcolor}

\usepackage{times,latexsym}
\usepackage{url}
\usepackage[T1]{fontenc}

\usepackage[acceptedWithA]{tacl2021v1}
\usepackage{xspace,mfirstuc,tabulary}

\usepackage{booktabs}
\usepackage{inconsolata}

\usepackage[T1]{fontenc}

\usepackage[utf8]{inputenc}

\usepackage{microtype}

\usepackage{amsmath}
\usepackage{amssymb}
\usepackage{graphicx}
\usepackage{float}
\usepackage{ntheorem}
\usepackage{listings}
\usepackage{xspace}

\newcommand{\dllmath}{\Delta}

\newcommand{\fphi}{f_{\boldsymbol{\phi}}}
\newcommand{\dll}{$\dllmath$\xspace}
\newcommand{\ent}{\mathrm{H}}

\newcommand{\bx}{\mathbf{x}}
\newcommand{\bxtarget}{\bx^{\mathrm{tgt}}}
\newcommand{\bxbase}{\bx^{\mathrm{base}}}
\newcommand{\words}{\boldsymbol{w}}

\newcommand{\vocab}{\overline{\Sigma}}
\newcommand{\eos}{\ensuremath{\textsc{eos}}\xspace}
\newcommand{\defeq}[0]{\mathrel{\stackrel{\textnormal{\tiny def}}{=}}}

\newcommand{\defn}[1]{\textbf{#1}}

\DeclareMathOperator*{\expect}{\mathbb{E}}

\newtheorem{defhypothesis}{Hypothesis}

\usepackage[textsize=tiny]{todonotes}

\title{Testing the Predictions of Surprisal Theory in 11 Languages}

\newcommand{\ucambridge}{2}
\newcommand{\ethz}{1}
\newcommand{\mituni}{3}

\author{
 Ethan G. Wilcox$^{\ethz}$~
 Tiago Pimentel$^{\ucambridge}$~
 Clara Meister$^{\ethz}$~
 \textbf{Ryan Cotterell}$^{\ethz}$~
 \textbf{Roger P. Levy}$^{\mituni}$ \\
 $^{\ethz}$ETH Zürich~\;~  $^{\ucambridge}$University of Cambridge~\;~  $^{\mituni}$MIT\\
 \texttt{\href{mailto:ethan.wilcox@inf.ethz.ch}{ethan.wilcox@inf.ethz.ch}}~\;~
 \texttt{\href{mailto:tp472@cam.ac.uk}{tp472@cam.ac.uk}}~\;~ 
 \texttt{\href{mailto:clara.meister@inf.ethz.ch}{clara.meister@inf.ethz.ch}} \\
 \texttt{\href{mailto:ryan.cotterell@inf.ethz.ch}{ryan.cotterell@inf.ethz.ch}}~\;~  
 \texttt{\href{mailto:rplevy@mit.edu}{rplevy@mit.edu}}
}

\begin{document}
\maketitle
\begin{abstract}


Surprisal theory \citep{hale2001probabilistic, levy2008expectation} posits that less predictable words should take more time to process, with word predictability quantified as surprisal, i.e., negative log probability in context.
While evidence supporting the predictions of surprisal theory has been replicated widely, much of it has focused on a very narrow slice of data: native English speakers reading English texts. Indeed, no comprehensive multilingual analysis exists. We address this gap in the current literature by investigating the relationship between surprisal and reading times in eleven different languages, distributed across five language families. Deriving estimates from language models trained on monolingual and multilingual corpora, we test three predictions associated with surprisal theory: (i) whether surprisal is predictive of reading times, (ii) whether expected surprisal, i.e., contextual entropy, is predictive of reading times, (iii) and whether the linking function between surprisal and reading times is linear. We find that all three predictions are borne out crosslinguistically. By focusing on a more diverse set of languages, we argue that these results offer the most robust link to date between information theory and incremental language processing across languages.\looseness=-1
\end{abstract}

\section{Introduction}

Language processing is incremental and dynamic: When a reader encounters a word, they allocate a certain amount of time to process it before moving on to the next one. 
One influential theory for the mechanism underlying this process is surprisal theory \citep{hale2001probabilistic, levy2008expectation}, which states that the time required to successfully comprehend a word is based on its predictability. Notably, predictability is often quantified as \defn{surprisal} (negative log-probability given preceding context), from which the theory's name is derived. 
Suprisal theory is supported, empirically, by a number of studies which have found that surprisal is strongly correlated with psychometric measurements in large naturalistic reading corpora \citep[\textit{inter alia}]{demberg-keller:2008, wilcox2020predictive, shain-2019-large,shain-2021-cdrnn,meister2021revisiting,pimentel2022effect,hoover2022plausibility}. 
Put differently, a word's surprisal is a strong correlate of its processing effort, operationalized as reading time.\looseness=-1

However, there is one serious limitation with most previous studies: While making general claims about human language processing, they predominantly investigate reading times in \emph{English}. 
And, while a few studies have investigated surprisal effects in languages other than English, e.g., \citet{meister2021revisiting} in Dutch and \citet{kuribayashi2021lower, kuribayashi2022context} in Japanese, no systematic, crosslinguistic analysis has been performed. 
As multiple sentence processing phenomena exhibit significant crosslinguistic variation \citep{hillert1998sentence}, the extent to which surprisal theory generalizes crosslinguistically is a nontrivial limitation of the current state of the literature.\looseness=-1

In addition, two recent contributions which we discuss here have posited several extensions to surprisal theory---most influentially, (a) that contextual entropy, i.e., expected surprisal, also correlates with reading times, and (b) that the relationship between surprisal and reading time is \emph{linear} \citep{smith2013effect, wilcox2020predictive, shain2022large}.
Regarding (a), \citet{pimentel2022effect} and \citet{cevoli2022prediction} have argued for what may be considered an expanded version of surprisal theory where processing difficulty is still determined by surprisal, but where people's reading behavior is additionally sensitive to expected surprisal (contextual entropy). 
Building off prior work that has investigated the role of entropy in language processing \citep{hale2003information, roark2009deriving, linzen2016uncertainty, van2017approximations}, these recent studies suggest that readers may allocate reading times in advance of encountering a word, based on their expectations of how difficult the word will be to process. 
Regarding (b), a number of studies have found evidence that the linking function between reading times and surprisal is linear \citep{smith2013effect, wilcox2020predictive, shain2022large}. 
However, these results have been challenged recently, with different studies coming to different conclusions about the most appropriate linking function. 
In the past two years, for example, investigations have concluded that this function is sublinear \citep{brothers2021word}, linear \citep{shain2022large}, and superlinear \citep{meister2021revisiting, hoover2022plausibility}.
Here, we will use the term \defn{surprisal theory} to refer to both the core hypothesis that reading times are correlated with surprisal, as well as the two extensions---(a) and (b)---described above.

We address a gap in the current literature by investigating the predictions of surprisal theory, on eleven languages distributed across five language families.\footnote{Our languages (and families) are: Korean (Koreanic), Turkish (Turkic), Hebrew (Semitic), Finnish (Uralic), Dutch, English, German, Greek, Italian, Russian and Spanish (Indo-European).\looseness=-1
}
We enumerate these three predictions as hypotheses below.
\begin{defhypothesis} \label{hyp:surp-theory}
\defn{(Surprisal Hypothesis)} Surprisal is predictive of reading times.\looseness=-1
\end{defhypothesis}

\begin{defhypothesis} \label{hyp:entropy}
\defn{(Contextual Entropy Hypothesis)} Contextual entropy is predictive of reading times.\looseness=-1
\end{defhypothesis}

\begin{defhypothesis}
\defn{(Linear Link Hypothesis)} The linking function between surprisal and reading times is linear.\looseness=-1
\end{defhypothesis}

We facilitate crosslinguistic comparison by using the MECO dataset \citep{siegelman2022expanding}, which presents eye-tracking data on reading materials with the same content in each language. 
We estimate surprisal and contextual entropy from two types of autoregressive language models---a single, large, multilingual model (mGPT; \citealt{shliazhko2022mgpt}), as well as monolingual models trained on large and small datasets, where the small dataset is the same size across languages ($\approx 30$ million words). We quantify the psychometric predictive power of surprisal and contextual entropy (i.e., how well each predicts reading times) by including them as variables in linear regression models. These models are then trained to predict by-word reading times; if the log-likelihood of the regression improves after including these variables, we take this as evidence that those variables have psychometric predictive power \citep{frank2011insensitivity, fossum2012sequential, goodkind2018predictive}.\looseness=-1

We find that, in all languages tested, regression models that include surprisal are significantly better predictors of reading times over baselines which do not include surprisal, confirming the surprisal hypothesis. Additionally, we find that models which include contextual entropy are even better predictors of reading times in most languages tested, confirming the contextual entropy hypothesis. Finally, compatible with the linear link hypothesis, we find that models constrained to a linear relationship between surprisal and reading times are just as good as those that can express more complex relationships.
Overall, our results provide the largest crosslinguistic analysis of the relationship between reading and word-level information theoretic properties to-date.\looseness=-1

\section{Psycholinguistic Predictive Power}

Our behavior of interest is how long readers spend visually attending to a given word $w_t$ in its linguistic context, i.e., $w_t$'s reading time.
This quantity offers a window into the psychological processes that underlie language comprehension and is typically taken as a direct reflection of the word's processing difficulty \citep{rayner1998eye}.
A word's reading time can be measured via multiple experimental modalities, including self-paced reading (\citealp{jegerski2013self}) and the maze task \citep{forster2009maze, boyce2020maze}. In this work, we focus on eye-tracking measurements. These measurements have high temporal resolution and exhibit smaller spillover effects than self-paced reading \citep{smith2013effect}, where spillover is the effect of a word's properties on later words' reading behavior.

Following previous work investigating reading, we ask what factors associated with each word are helpful for predicting its reading times. 
In the following section,
we use the following notation.
With $w$, we denote a word taken from an alphabet $\Sigma$.
With $\words \in \Sigma^*$, we denote a string of words over the alphabet $\Sigma$.
We write $w_t$ for the word at index $t$ in a string $\words = w_1 \cdots w_T$ with $1 \leq t \leq T$.
Additionally, let $\eos \not\in \Sigma$ be a distinguished end-of-string symbol \emph{not} in $\Sigma$ and let $\overline{\Sigma} \defeq \Sigma \cup \{\eos\}$ be an augmented alphabet that includes $\eos$. 
With each word $w_t$ in a context $\words_{<t}$, we associate a real column vector of predictor variables $\bx_t$ that we believe may impact reading times.
Many of these predictors are attributes of $w_t$ itself, e.g., $w_t$'s length.
We use $\bx_t$ as predictors in a regression model $\fphi$ with
parameters $\boldsymbol{\phi}$.
The regression model is
estimated to predict $w_t$'s reading time from data.
In symbols, we write that
\begin{align}
   y(w_t, \words_{<t}) \sim \fphi(\cdot \mid \bx_t )
\end{align}
where $y(w_t, \words_{<t})$ is the reading time of word $w_t$ in context $\words_{<t}$.
To be explicit, in our formulation we treat reading times as a continuous quantity and, thus, $\fphi$ is a probability \emph{density}.\looseness=-1

In order to contrast different theories of language processing, we compare regression models with different vectors of predictor variables $\bx$ and with different architectures $\fphi$, each of which is taken to instantiate a different hypothesis about what underlying factors determine reading times.
We fit each regression model on a portion of our dataset and evaluate it by measuring the log-likelihood that it assigns to held-out data.
Models that lead to higher log-likelihood can be said to have better predictive power or psychological accuracy for human reading---and their associated theories are then taken to be better models of the underlying psycholinguistic processes \citep{frank2011insensitivity, fossum2012sequential}. 

Typically, for each experiment we will define a \defn{target regression model}, which is trained to predict the reading times of individual words from a set of baseline predictors plus a predictor of interest (e.g., surprisal or contextual entropy). 
For a specific index $t$, we will refer to these predictors as our \defn{target predictors} and denote them as $\bxtarget_t$.
We also define a \defn{baseline regression model} that includes only the \defn{baseline predictors}, which are a subvector of the target predictors, denoted as $\bxbase_t$ for a specific $t$.
We denote baseline and target regression models symbolically as $\fphi(\cdot \mid \bxtarget_t)$ and $\fphi(\cdot \mid \bxbase_t)$, respectively. 
Unless otherwise specified, the regression models that we use in this study are all linear. 
The choice to use linear linking functions, and whether this assumption is warranted, is addressed directly in Section \ref{sec:rt-surp-link}. In order to assess whether the target predictors have contributed to better predictive power, we will inspect the (average) by-word difference in log-likelihood assigned by the two regression models to a held-out dataset \citep{goodkind2018predictive, wilcox2020predictive}. 
Following previous studies, we refer to this metric as the \defn{delta log-likelihood} $\Delta$, which
is defined, for a specific index $t$, as\looseness=-1
\begin{equation}
\begin{split}
    \Delta_t = \log &\fphi\left(y(w_t, \words_{<t}) \mid \bxtarget_t\right)\\
    &- \log \fphi\left(y(w_t, \words_{<t}) \mid \bxbase_t\right) 
\end{split}
\end{equation}
where $y(w_t, \words_{<t})$ is the observed reading time of word $w_t$ in context $\words_{<t}$.
The complete metric $\Delta$ is the average of $\Delta_t$ over all word indices.
A positive \dll means that the target predictors contribute to psycholinguistic predictive power above the baseline predictor, whereas a \dll of zero indicates that the added predictors either lack a robust relationship with reading times or that their functional relationship cannot be approximated by the class of models $\fphi$ we employ.\footnote{In practice, negative values of \dll are also possible; they indicate overfitting, and imply the same theoretical conclusion as a \dll of 0.\looseness=-1
} Below, we briefly introduce the two target predictors associated with the theories that we wish to test: surprisal and contextual entropy.

\subsection{Surprisal}
\newcommand{\bw}{\boldsymbol{w}}
\newcommand{\bW}{\boldsymbol{W}}


The surprisal \citep{shannon1948mathematical} of a word $w_t$ measures the information content it conveys in the context in which it appears.
Using Shannon's formulation of entropy, we can define surprisal as
\begin{align}
    s_t(w_t) \defeq - \log_2 p(w_t \mid \bw_{<t} ) 
\end{align}
where $p(\cdot \mid \bw_{<t} )$ is the true distribution over words $w \in \vocab$ in context $\bw_{<t}$, which we omit from the notation for brevity.
We focus here on reading, where the relevant context to compute surprisal is the $w_t$'s preceding words $\bw_{<t}$. 
However, in our studies, we do not have access to the true distribution $p(\cdot \mid\bw_{<t})$ and instead estimate it using an autoregressive language model, as is common in previous studies \citep{smith2013effect, goodkind2018predictive, wilcox2020predictive}.


\subsection{Contextual Entropy}

The contextual entropy of a $\vocab$-valued random variable $W_t$ at index $t$ is the expected value of its surprisal, which can be expressed as
\begin{subequations}
\begin{align}\label{eq:shannon_ent}
    \ent(&W_t \mid \bW_{<t} = \bw_{<t}) \defeq \expect_{w \sim p(\cdot \mid \bw_{<t})} [s_t(w)] \\
    &= - \sum_{w \in \vocab} p(w \mid \bw_{<t}) \log_2 p (w \mid \bw_{<t}) \
\end{align}
\end{subequations}
Again, as we do not have access to the true distribution $p$, so we resort to estimating the contextual entropy using an autoregressive language model. 

Prior work has investigated the relationship between different contextual entropy and reading behavior: A number of studies have investigated entropy reduction, or the extent to which $w_t$ reduces uncertainty over possible next words \citep{frank-2010-uncertainty, frank2013uncertainty} or the possible incremental parses that can be assigned to a sentence prefix \citep{hale2003information, hale2006uncertainty}.
Other researchers have investigated the effect of successor entropy, i.e., the entropy of $W_{t+1}$, on predicting the current-word reading times \citep{roark2009deriving,linzen2016uncertainty,van2017approximations}.\footnote{When computing $W_{t+1}$, it is common to treat $W_t = w_t$ as observed.\looseness=-1}
In contrast, we look at the effect of $W_t$'s contextual entropy on prediction, following \citet{pimentel2022effect} and \citet{cevoli2022prediction}.
As discussed in \citet{pimentel2022effect}, investigating contextual entropy separately from surprisal can uncover to what extent reading behavior is responsive (i.e., driven by surprisal) or anticipatory (i.e., driven by expected surprisal). 
\citet{pimentel2022effect} specifically found that contextual entropy is a significant predictor of reading times on 3 out of 4 of their tested English eye-tracking and self-paced reading datasets.\looseness=-1

\section{Experimental Setup}

\subsection{Dataset} \label{sec:dataset}

We use the Multilingual Eye Movement Corpus \citep[MECO;][]{siegelman2022expanding}. MECO contains eye-tracking data from L1 speakers (between 29 and 54 per language) for 12 simplified Wikipedia-style articles in thirteen languages; these languages are from five different language families. 
Articles in the MECO corpus went through an iterative translation process by separate teams of translators to ensure that article content was the same across languages and range from a minimum 1,487 total words (Finnish) to a maximum 3,021 total words (Russian). 
The eleven languages we include in our analysis are: Korean (Koreanic), Turkish (Turkic), Hebrew (Semitic), Finnish (Uralic), Dutch, English, German, Greek, Italian, Russian, and Spanish (Indo-European).\footnote{The dataset also includes Norwegian and Estonian, however these are not supported by our multilingual language model and therefore excluded.} 
While this sample is still biased towards Indo-European languages, it is more diverse than other previous studies, which have tended to focus exclusively on a single language. \looseness=-1

The following pre-processing steps were taken: Words that were skipped on the first pass were given a reading-time of zero and included in the analysis. 
Eye-tracking datasets report multiple different word-based measurements of reading times, of which we use three \citep{rayner1998eye}: The \defn{first fixation} is the duration of the first fixation on a word during its first pass. 
\defn{Gaze duration} is the sum of all first-pass fixations on a word. And \defn{total fixation} time is the sum of all fixations on a word during the trial. While we report results for all three for the sake of completeness, our discussion will focus on results for gaze duration as has been done in previous studies, e.g., \citealt{wilcox2020predictive}. 
First fixation times are typically associated word identification \citep{clifton2007eye} and are expected to not reflect strong contextual influences. Total reading durations can be influenced by material from the right context (i.e., regressive saccades). Thus, for studies that focused on progressive movement through a text, such as ours, gaze duration is expected to be most strongly associated with first-pass processing difficulty, which is our cognitive process of interest. 
For each of these metrics, we fit a regression model on averages of the reading time measures taken across subjects, as has been done in previous work \citep{smith2013effect, wilcox2020predictive}. This step was performed to mitigate the potentially high by-participant variance present in eye-tracking data.\looseness=-1

\subsection{Language Models} \label{sec:models}

\begin{table}[]
    \centering
    \small
    \begin{tabular}{lcc}
    \toprule
        Language & Code & \# Training Tokens (mil)  \\
    \midrule
        Dutch & \texttt{du} & \phantom{0,}171 \\
        English & \texttt{en} & 1,966 \\
        Finnish & \texttt{fi} & \phantom{0,0}89 \\
        German & \texttt{ge} & \phantom{0,}883 \\
        Greek & \texttt{gr} & \phantom{0,0}57 \\
        Hebrew & \texttt{he} & \phantom{0,}112 \\
        Italian & \texttt{it} & \phantom{0,}376 \\
        Korean & \texttt{ko} &  \phantom{0,0}75 \\
        Russian & \texttt{ru} & \phantom{0,}488 \\
        Spanish & \texttt{sp} & \phantom{0,}508 \\
        Turkish & \texttt{tr} & \phantom{0,0}48 \\
    \bottomrule
    \end{tabular}
    \caption{Training data information for our monolingual transformer models, noted as monoT(all)}
    \label{tab:monot_dataset}
\end{table}

We derive surprisal and contextual entropy estimates from both monolingual and multilingual models, which we describe in greater detail below.

\paragraph{Monolingual Models} We train monolingual transformer models using the Wiki40B dataset \citep{guo-etal-2020-wiki}, from which we rely on the training and validation splits from the original paper for each of our analyzed languages.
We first fit language-specific UnigramLM tokenizers \citep{kudo-2018-subword} with a vocabulary size of 32k on the training portion of this dataset, which we then use to tokenize both the Wiki40B and MECO text into subword units.
We then train two models per language, with different amounts of training data:
For the \defn{monoT(all)} variant, we train the model on the total amount of data in Wiki40B for each language; for the \defn{monoT(30m)} variant, we subsample $\approx$ 30 million tokens from each language. For a list of the training dataset sizes for the monoT(all) models, as well as a list of language codes that will be used in figures, see Table \ref{tab:monot_dataset}.
We train all our models using fairseq \citep{ott2019fairseq}, following their recommended language modeling training hyper-parameters. We use a standard decoder-only transformer with 6 layers, a context window size of 512 tokens, and shared input--output embeddings. We train our models using Adam \citep{kingma2017adam}, with a learning rate of $5e^{-4}$, 4000 warm-up updates, and dropout of 0.1.
For both of our monolingual models, as well as the multilingual model described below, per-word surprisals are computed by summing over subword unit surprisals, which is the appropriate procedure since surprisal decomposes additively over the units compromising a signal.
Because of spurious ambiguity inherent in the tokenization scheme, an efficient algorithm to estimate contextual entropy over full words is unavailable to us; such an algorithm requires summing over an infinite number of sub-word combinations. 
Instead, we simplify this computation by estimating contextual entropy over one single step of sub-word tokens as suggested in \citet{pimentel2022effect}. 
Techniques similar to this have been employed previously in studies of entropy \citep{frank-2010-uncertainty}, e.g., account for clitics and contractions.\looseness=-1

\paragraph{Multilingual Model} We use mGPT \citep{shliazhko2022mgpt}, a multilingual autoregressive language model, which was trained with the GPT-3 architecture on 60GB of text\footnote{\citet{shliazhko2022mgpt} report that their combined dataset contains 489 billion characters. Assuming a crosslinguistic average of $\approx$ 5 characters per word, this puts their training set at slightly under 100 billion words.} from a combination of Wikipedia and the Cleaned Common Crawl Corpus \citep{raffel2020exploring}.\looseness=-1

\paragraph{Context Length}
One recent study has hypothesized that, when deriving surprisal estimates for psycholinguistic modeling, the size of the context window can bias estimates \citep{hoover2022plausibility}. Their reasoning is that short context windows could shift probability mass away from very low-frequency words, which would be better predicted from longer contexts. Therefore, we estimate surprisal and contextual entropy from mGPT in two contexts: In short contexts the model is given only the current sentence (up until the current word); in long contexts we use the model's full input window size of 512 tokens.
We use long contexts for our first analysis, and use both contexts for our second analysis, which investigates both the shape of the reading times--surprisal linking function and the influence of context length on these results.\looseness=-1

\begin{figure*}
    \centering
    \begin{minipage}{\textwidth}
    \centering
    \includegraphics[width=\textwidth]{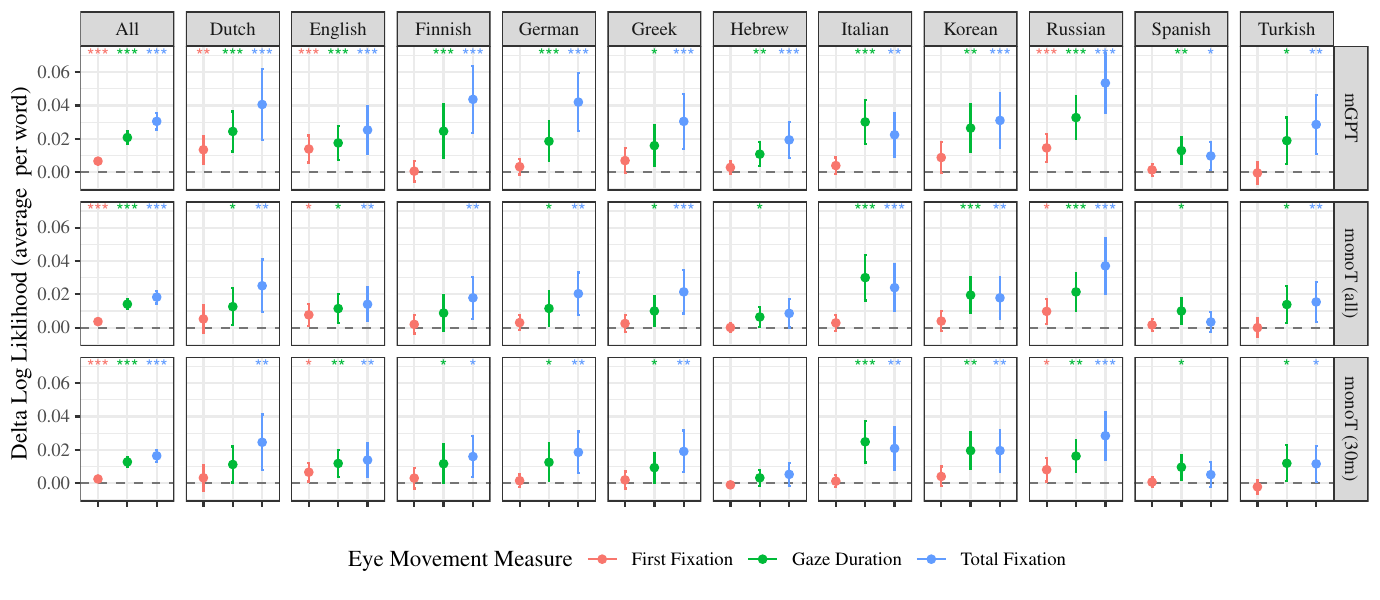}
    \end{minipage}
    \vspace{-15pt}
    \caption{\textbf{Predictive Power of Surprisal Across Languages:} Positive values mean surprisal contributes to predicting the reading times over a baseline where surprisal is removed. Error bars indicate 95\% confidence intervals. Stars indicate the significance of a paired permutation test. 
    We find a consistent significant effect of surprisal across languages for language models that are both multilingual (top row) and monolingual (bottom two rows), and for both progressive gaze duration and total fixation.
    }
    \label{fig:surprisal}
\end{figure*}

\paragraph{Psychological Plausibility} Increasingly, researchers that use language models for cognitive modeling have considered their psychological plausibility as estimates of humans' internal notions of word predictability. In particular, some researchers have compared the size of the models' training data to the amount of linguistic experience of the average human child \citep{zhang-etal-2021-need}. Assuming that children are typically exposed to $\approx 11$ million words per year as an upper limit \citep{hart1995meaningful}, then the mGPT model is trained on multiple human lifetimes' worth of language data.
The monoT(all) models are trained on data scales equivalent to or less than one human lifetime,\footnote{The only exception is English, which at $\approx 2$ billion words is about two lifetime's worth of linguistic data, assuming the 11-million word per year estimate of \citet{hart1995meaningful}.} and the monoT(30m) models are trained on data equivalent to the linguistic exposure of a young child. 
However, we argue that the psychological plausibility of a model's next-word predictions is not completely determined by whether that model's training data is the same size as the amount of data a human learner is exposed to.
Indeed, there is a body of evidence suggesting that, beyond a certain minimal amount of data, the more data a model is trained on, the more human-like that model's next-word predictions become \citep{goodkind2018predictive, wilcox2020predictive}. 
All of our models are trained of an amount of data within this range.
However, at the other end of the scale, the relationship flips: Models trained on an extremely large amount of data seem to be slightly \emph{worse} predictors of human reading \citep{shain2022large, oh2022does}.
For our models, training datasets are uni-modal (i.e., language only) and learning is with arguably weaker priors for language-like structure, whereas humans learn from multi-modal data with potentially much stronger priors for linguistic structures.
Likely, more data makes up for the lack of multi-modal data and uninformative priors.\looseness=-1

\subsection{Regression Models} \label{sec:regression_models}

All of our regression models are fit to predict the reading time $y(w_t, \words_{<t})$ of a word $w_t$ in a context $\words_{<t}$ from the predictor vector $\bx_t$.
In addition to looking at the word $w_t$, our predictor includes quantities derived from the previous two words $w_{t-1}, w_{t-2}$ to control for potential spillover effects.
We will refer to the three words $w_t,w_{t-1},w_{t-2}$ as our \defn{regressor words}. Following previous work in this area, all regression models include the word length and negative log frequency (i.e., unigram surprisal), as estimated by \citet{speer2022wordfreq}, for all regressor words in a predictor $\bx_t$ for a specific index $t$.
The predictors above constitute our (context invariant) baseline predictors. 
Regression models are trained and evaluated using 10-fold cross validation.
For more information on the regressions used in each of our experiments, see Appendix \ref{app:regressions}.
The significance of the observed \dll values between target and baseline models is assessed via a paired permutation test that checks whether \dll is significantly different from zero. 
We use permutation tests because our comparisons because they make no assumption about the distribution of the test statistic. Instead, the test uses the empirical distribution of differences in likelihoods, as estimated using averages computed over permutations of likelihoods, in order to compute $p$-values.\looseness=-1

\section{Results}
\subsection{Surprisal} \label{sec:surp-result}

To test the surprisal hypothesis, we fit a target regression model whose predictors includes the surprisals of our regressor words plus our baseline predictors described above.
We compare this to a baseline that does not include the surprisal predictors. 
For this and subsequent tests, we calculate results for each language individually, as well as for the combined data from all languages. Results can be seen in Figure \ref{fig:surprisal} broken down by language, model, and each of our three word-based measurements of reading time. We observe a clear pattern in the results across the languages: Positive \dll in nearly every test for gaze duration and total fixation, and less consistently positive \dll for first fixation, where, as noted before, we would not necessarily expect surprisal effects to show up. 
Looking at the results for each model, we observe the most robust results for mGPT, where \dll is significantly greater than zero in every language for gaze duration and total fixation. For the monolingual models, we observe more robust effects for the monoT(all) model over the monoT(30m) model, which is sensible given the latter's limited training data size.\looseness=-1


\begin{figure*}
    \begin{minipage}{\textwidth}
        \centering
\includegraphics[width=\textwidth]{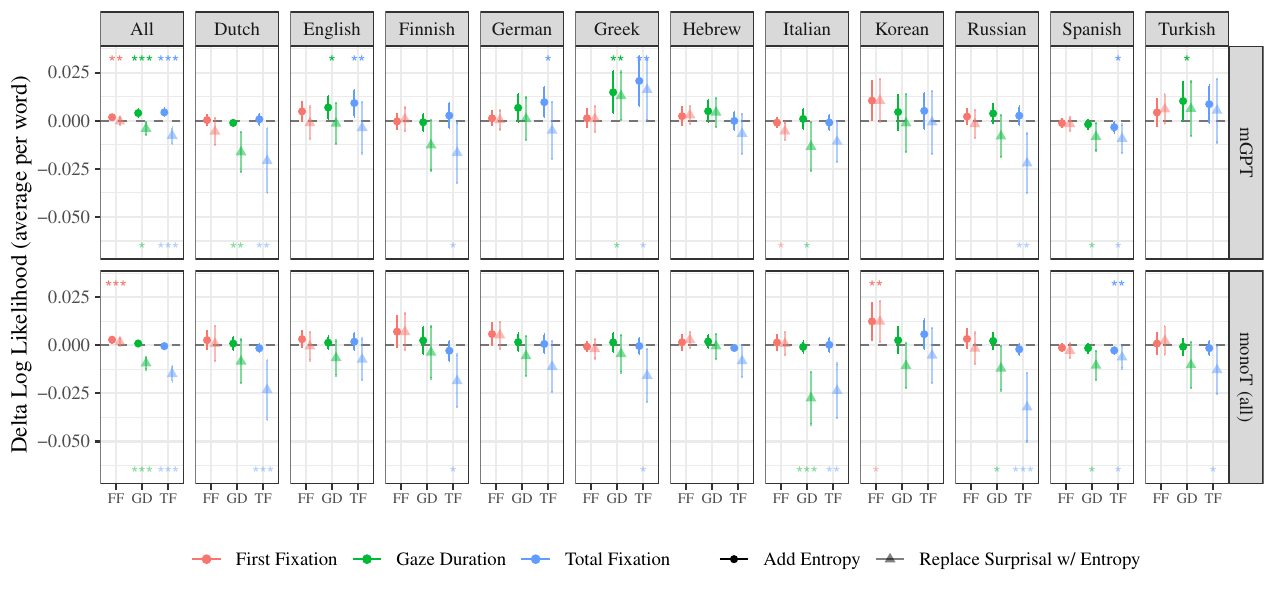}
    \end{minipage}
        \vspace{-15pt}
    \caption{\textbf{Psychometric Predictive Power of Contextual Entropy Across Languages:} Positive values mean contextual entropy contributes to predicting the reading times of $w_t$. Error bars are 95\% confidence intervals across the 10 folds of held-out data. Stars indicate the significance of a paired permutation test. We find that replacing surprisal with entropy tends to hurt predictive power, while adding entropy tends to help.}
    \label{fig:entropy}
\end{figure*}

For an aggregate test of the effects of surprisal, we fit an additional regression model on the combined data from all languages to predict gaze duration with random by-language effects. We use a fully maximal random effect structure, as advocated in \citet{barr2013random}.
We find that the model with surprisal leads to significantly greater than zero \dll in all cases ($p<0.001$). 
Although surprisal leads to a positive \dll across languages, we do observe some variation in the magnitude of this effect, or the predictive power obtained by regression model. For both mGPT and monoT(all) we observe the highest predictive power in Russian and Dutch, with lower predictive power in Spanish, English, and Hebrew. 
One natural question to ask is whether imbalances in the model's training data leads to some of this variation---do models make better predictions for language where they have seen more data?
However, there are converging pieces of evidence from our data suggesting that differences in dataset size is \emph{not} the main cause of the by-language variation. 
First, both mGPT and monoT(all) show relatively lower predictive power for some large-data languages such as Spanish. 
Second, and quite interestingly, similar patterns of predictive power can be observed for our monoT(30m) models, where training dataset size is controlled across languages. 
Here, as with the other models, we observe larger values of \dll in Italian and Russian and smaller values of \dll in Spanish and Hebrew. 
These results pose a puzzle, as the languages for which the models obtain higher \dll are not obviously different from those for which the models obtain lower \dll, in terms of their linguistic features.
For example, Spanish (lower \dll) and Italian (higher \dll) are both Romance languages. 
Further investigation is needed to determine if these patterns hold up for other crosslinguistic reading time datasets.

\subsection{Contextual Entropy} \label{sec:entropy}

To test the contextual entropy hypothesis we first fit a single baseline regression model.
Our baseline regression model includes the surprisal of all regressor words, plus baseline predictors. We then evaluate target regression models in two variants: For the \textit{replace} regression model, we replace surprisal with contextual entropy for all regressor words. For the \textit{add} regression model, we add an additional term of contextual entropy for all regressor words. As results do not change much between our monolingual language models, we present results for monoT(all).\looseness=-1

Results can be seen in Figure \ref{fig:entropy}, where the replace regression is indicated with a triangle and the add regression is indicated with a circle. First, we find that replacing surprisal with entropy tends to hurt predictive power in most cases. For example, for mGPT, gaze duration \dll is negative in 7/11 languages and significantly so in three (Spanish $p<0.05$, Dutch ($p<0.01$) and Italian ($p<0.05$)), implying overfitting. 
Negative effects are even stronger for the monoT(all) model, where we find negative gaze duration \dll in every language except Hebrew (results are significant in 3/11). 
Adding entropy as an \emph{additional} predictor, on the other hand, generally improves the model's predictive power. 
For example, for mGPT and gaze duration, \dll from the add regression is positive in 8/11 languages, and significantly so in 3 (English, Greek and Turkish). 
In addition, \dll is significantly positive for the add regression for all three reading time measures when data is combined across languages, as shown in the `All' column at the left of Figure \ref{fig:entropy}. Results are less strong for monoT(all), where positive \dll shows up predominantly for first fixation. As before, we run an aggregate test with data from all languages including by-language random effects.\footnote{Following the same methodology as the previous test, we look at the effect of adding or replacing surprisal across all regressor words.} 
For gaze duration, we find that adding contextual entropy leads to positive \dll (mGPT, $p<0.001$; monoT(all), $p<0.01$) and that replacement leads to negative \dll (mGPT, $p<0.01$; monoT(all), $p<0.001$). 
Overall, we take these results as being in line with those reported in \citet{pimentel2022effect}.
Our findings suggest that contextual entropy has a weak---albeit consistent---effect on reading times across languages, and therefore that participants may be pre-planning their processing times based on the expected surprisal of upcoming words.

\subsection{Variation Across Languages} \label{sec:variation}

The crosslinguistic relationship between \dll and language model quality is relevant to current debates about about whether language models can plausibly be used to understand psycholinguistic processes. As mentioned in Section \ref{sec:models}, it has been observed that, within English, models with lower perplexity tend to exhibit better predictive power \citep{goodkind2018predictive, wilcox2020predictive}.
However, studies on Japanese have failed to replicate these results, suggesting that the relationship does not hold for all languages \citep{kuribayashi2021lower}. Further, \citet{oh2022does} and \citet{shain2022large} show that this relationship may not hold even in English for the most recent language models. 
To investigate this, we compute, for mGPT, the Pearson's correlation between \dll and test set perplexity, as reported in \citealt{shliazhko2022mgpt}, both across languages, as well as across language families.\footnote{For language families, \dll and perplexities are within-family averages.} 
For this analysis we show results only for mGPT only and leave a full analysis, comparing different monolingual models for future work.

\begin{figure}
    \begin{minipage}{\columnwidth}
        \centering
\includegraphics[width=\textwidth]{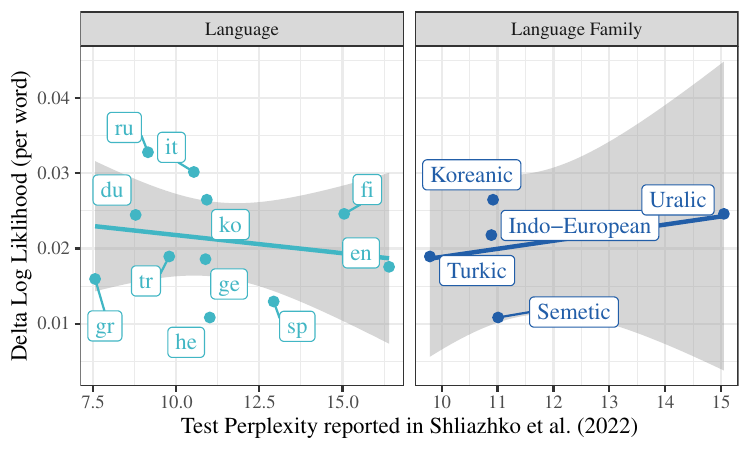}
    \end{minipage}
    \caption{\textbf{Test Perplexity versus \dll (mGPT):} We do not find a significant correlation between the \dll and mGPT's perplexity for a language or language family. }
    \label{fig:dll-ppl}
\end{figure}

The correlations can be seen in Figure \ref{fig:dll-ppl}. We do find a negative correlation across languages, however it is not significant ($\rho=-0.186, p=0.6$). We do not find any evidence of correlation in the language family data. Although the negative by-language correlation suggests that, for languages where mGPT has lower perplexity, it may be a better model of psycholinguistic behavior, the lack of significance is in line with the negative results from Japanese.

Notably, there are important differences between this analysis and the studies cited above, which train a number of different language models within a single language and a single shared vocabulary, as opposed to comparing the outputs of a single multilingual language model across languages as we do here. Additionally, although mGPT does share a single vocabulary across languages, different languages might be \textit{a priori} harder or easier to language-model \citep{cotterell2018all, mielke2019kind}, and quality of the tokenization might vary across languages as well. Thus, more fine-grained linguistic controls are necessary before making strong conclusions about the relationship between perplexity and psychometric predictive power across languages.

\begin{figure*}
    \begin{minipage}{\textwidth}
        \centering
\includegraphics[width=\textwidth]{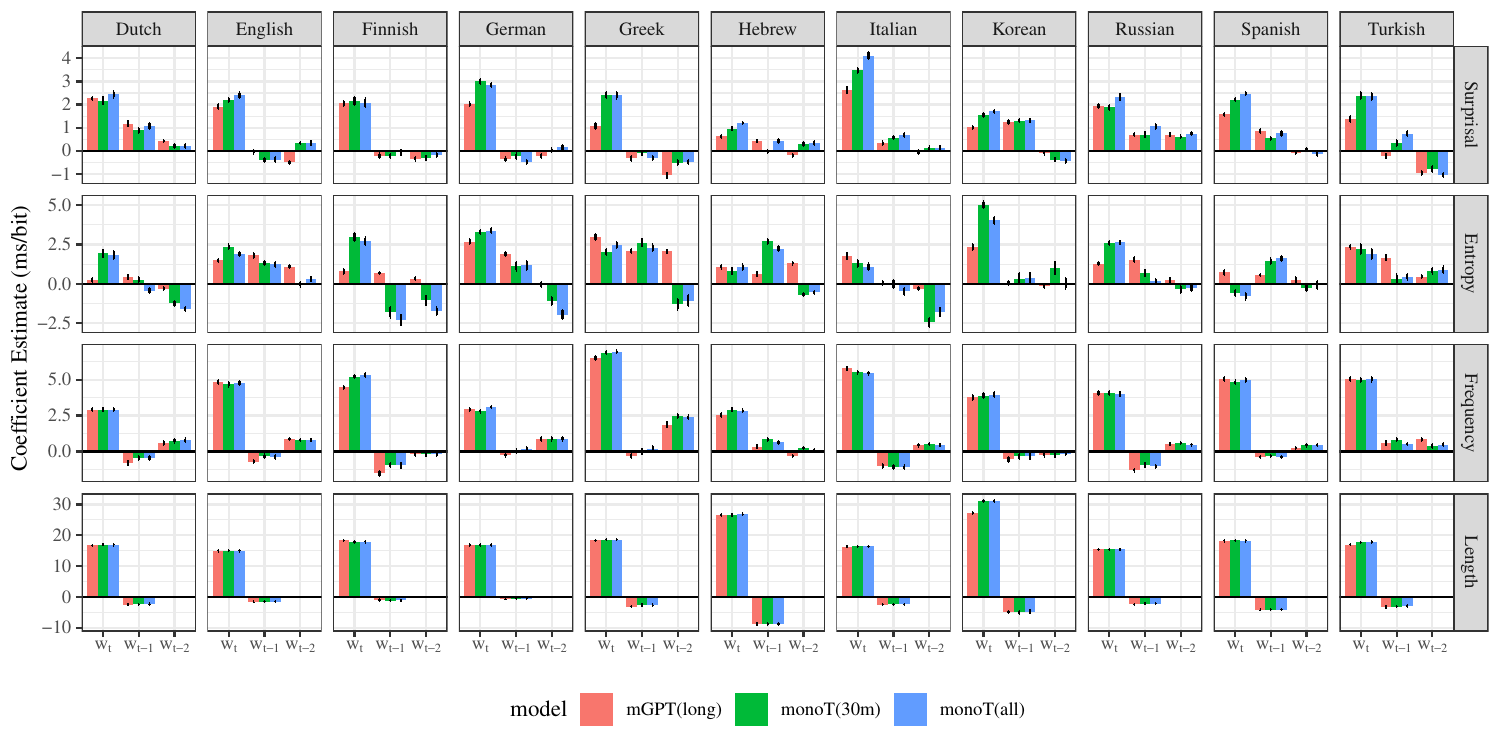}
    \end{minipage}
        \vspace{-15pt}
    \caption{\textbf{Model Coefficients:} Coefficients for a linear model that includes surprisal, entropy, frequency and length. Coefficients are shown for each regressor word individually. Zero is indicated with a black line and scales differ for each row. Error bars indicate 95\% CIs across folds of data. }
    \label{fig:coeffs}
\end{figure*}

\subsection{Model Coefficients} \label{sec:coeffs}

How do surprisal, entropy, frequency and length individually affect reading times? Figure \ref{fig:coeffs} shows the estimates for each of our predictor variables, estimated across 10 folds of data. Unlike the figures presented above, effects are broken down by the coefficients for each of our regressor words from $w_t$ (on the left of each facet) to $w_{t-2}$ (on the right of each facet). Note that effect size here does not correspond to the predictive power of the model as a whole, but rather the impact of word-level properties on reading times. Because predictor variables are not normalized, units are different across rows. The top two rows indicates the estimated slowdown in milliseconds for each additional bit (of surprisal or entropy). The second row indicates slowdown for each additional occurrence per billion words of text (on a log scale). And the bottom row indicates slowdown for each additional character in the word. \looseness=-1

We find a consistent effect of surprisal for $w_t$ of between 2-4 ms/bit. There is some inter-language variability, with the smallest effect for Hebrew, and larger effects for Dutch, Russian, Greek and Italian. We find smaller effects for $w_{t-1}$, ranging from between 0-2 ms/bit. There is no obvious effect of surprisal for $w_{t-2}$. Overall, these results differ slightly from those reported in \citet{smith2013effect}, who investigate reading times on the English Dundee Corpus \citep{kennedy2003dundee} and find a stronger effect for $w_{t-1}$ than we do. However, our results are not inconsistent with the relatively lower spillover effects traditionally observed in eye-tracking data.

Turning to contextual entropy, we find slightly smaller effects, and slightly more variance between languages. There is no obvious relationship between the effect sizes for surprisal and contextual entropy. For example, Dutch, which has a larger surprisal effect, has one of the smallest effect sizes for entropy. For negative log frequency (i.e., unigram surprisal), we find a consistently positive effect for $w_t$, as expected implying that as words get more frequent they take less time to read. For $w_{t-1}$ and $w_{t-2}$ effects are much smaller and less consistent across languages. For example, Dutch, Finnish, Italian and Russian all have consistently negative frequency effects for $w_{t-1}$, whereas in Turkish and Hebrew, these effects are positive.

We find consistent effects for word length, which are positive for every language on $w_t$. We also find consistent negative effects for $w_{t-1}$. This may be due to the fact that readers are likely to skip a word if it comes after a long word, which would be associated with a reading time of zero in our analysis. Overall, these coefficient estimates are in line with previous reading time studies and further highlight the crosslinguistic consistency of our results.

\section{Surprisal--RT Linking Function} \label{sec:rt-surp-link}

\begin{figure*}
    \begin{minipage}{\textwidth}
    \centering
    \includegraphics[width=\textwidth]{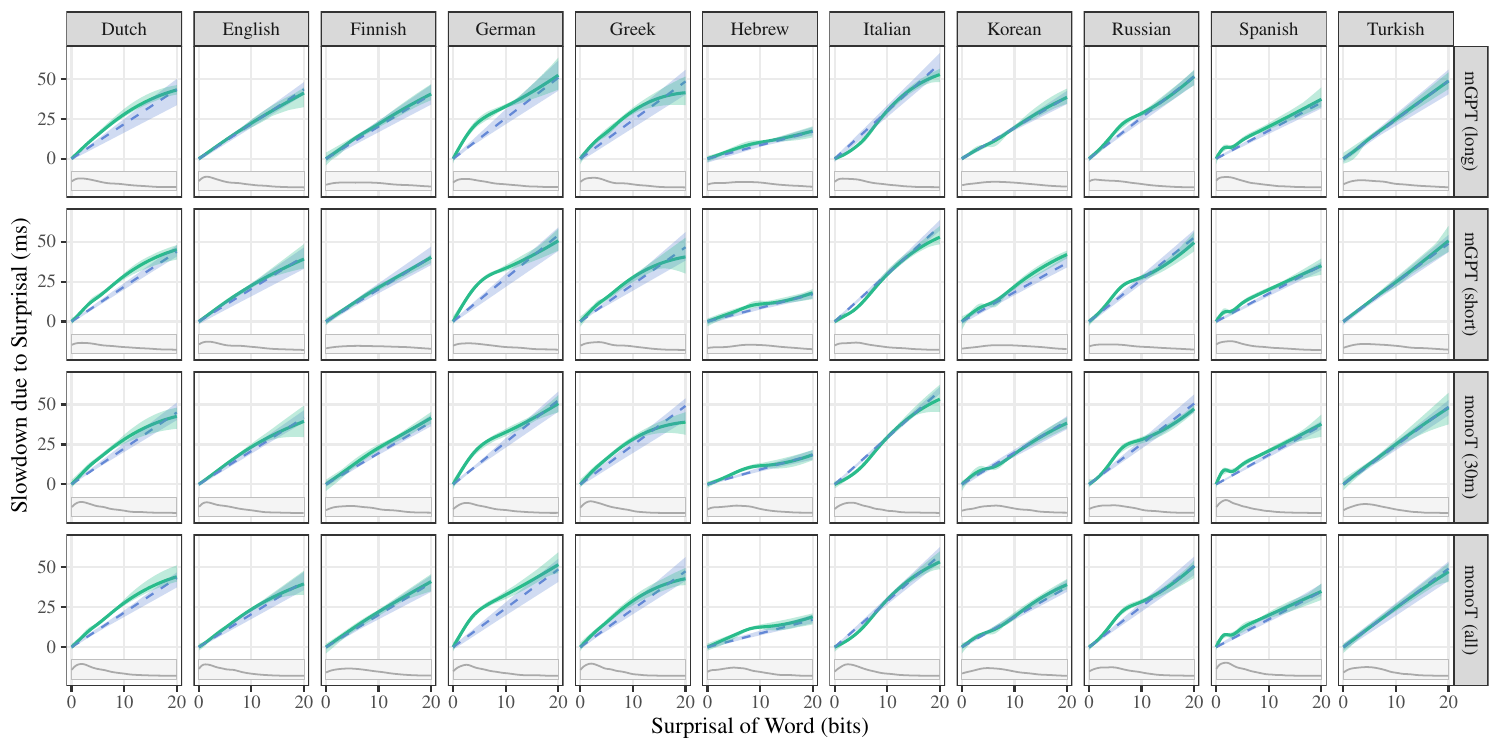}    \end{minipage}
    \caption{\textbf{Surprisal versus Reading Time Relationship:} Non-linear GAMs are in green while linear control GAMs are in dotted blue. Shaded regions represent bootstrapped 95\% confidence intervals. Results are for gaze duration. Grey subplots indicate the distribution of surprisal values. 
    We find that GAMs recover a linear relationship between surprisal and reading-time slowdown.\looseness=-1
    }
    \label{fig:gam-surp}
\end{figure*}

\noindent The regression models we have been using to assess \dll have implicitly assumed a linear linking function between surprisal and reading time---a relationship that has been empirically verified in some previous studies in English \citep{smith2013effect, wilcox2020predictive,shain2022large}. 
Other recent studies, however, have questioned linearity, including \citet{meister2021revisiting} and \citet{hoover2022plausibility}, who argue for a \emph{super}linear relationship, and \citet{brothers2021word}, who argue for a \emph{sub}linear relationship. 
In this section, we directly test the linear link hypothesis. We compare the \dll of our linear regression models against regression models that can capture non-linear relationships.
We present results exclusively for gaze duration for the reasons discussed in Section \ref{sec:dataset}. \looseness=-1

\subsection{Visualizing the Link with GAMs} \label{sec:gams}
In order to visualize the link between surprisal and reading times, we use generalized additive models (GAMs), a class of models that can fit non-linear relationships between predictor and response variables. 
Given the less-constrained hypothesis space of the GAM, if the model finds a relationship that is (visually) linear, this is good first evidence that the underlying effect is linear. We fit a GAM to predict reading times from word frequency, length and surprisal, derived for short contexts (sentence level) and long contexts (document level). We include smooth terms for current and previous word surprisal, as well as tensor product terms for a non-linear interaction between log-frequency and word length. 
By way of comparison, we also fit a GAM that enforces a linear effect of surprisal, following \cite{hoover2022plausibility}. For this comparison, we fit new models, all using the \texttt{mgcv} library, as opposed to simply comparing GAMs to our linear models from the previous section, to ensure that the effects of our baseline variables are exactly the same between models in this section.\footnote{For these analyses we choose to only include surprisal, frequency and length from $w_t$ and $w_{t-1}$ as predictors. This was done because of the minimal effects found on $w_{t-2}$ in our analysis of coefficients (see Figure \ref{fig:coeffs}). A sample GAM call for this analysis is given in Appendix \ref{app:regressions}.}
For each language and language model combination, we visualize the fitted curve using 10-fold cross validation, i.e., we train a GAM model on 9 of the 10 folds and sample reading times from the trained model using the remaining fold. 
To sample reading times, 
we vary the surprisal values for $w_t$ ranging 0--20 in increments of 0.1.
No other predictors are fed into the model.

The visualizations of the estimated GAMs for effects on $w_t$ can be seen in Figure \ref{fig:gam-surp}. 
Below the fit, we show density plots for surprisal values in the corpus. 
The results are consistent across languages and contexts. 
Visually, the non-linear GAMs capture the effect of surprisal on reading times by fitting an approximately linear curve, which sometimes falls directly on top of the linear control GAM (e.g., for Finnish and Turkish). 
Unlike \citet{hoover2022plausibility} we do not find a consistent difference for fits between surprisals derived in short contexts versus long contexts.
We note, however, that \citet{hoover2022plausibility} finds superlinear trends specially for their best examined models (e.g., GPT-3), which may outperform multilingual mGPT.\looseness=-1

\subsection{Testing Linearity}

Although the GAM fits in Figure \ref{fig:gam-surp} are \textit{visually} linear, we would like to test the question of linearity with a more rigorous method. To do so, we compare the \dll of the linear and non-linear GAMs described above. \dll is calculated by comparing each model to a shared baseline that includes only tensor product terms for frequency and length. The idea is that if the underlying relationship between surprisal and reading time is non-linear, then the non-linear GAMs should be able to achieve higher \dll, whereas if the underlying relationship is linear then the non-linear GAMs would not have an advantage. Thus, a consistently null result across languages suggests that the relationship is linear.

\begin{figure}
    \centering
    \begin{minipage}{0.48\textwidth}
    \includegraphics[width=\textwidth]{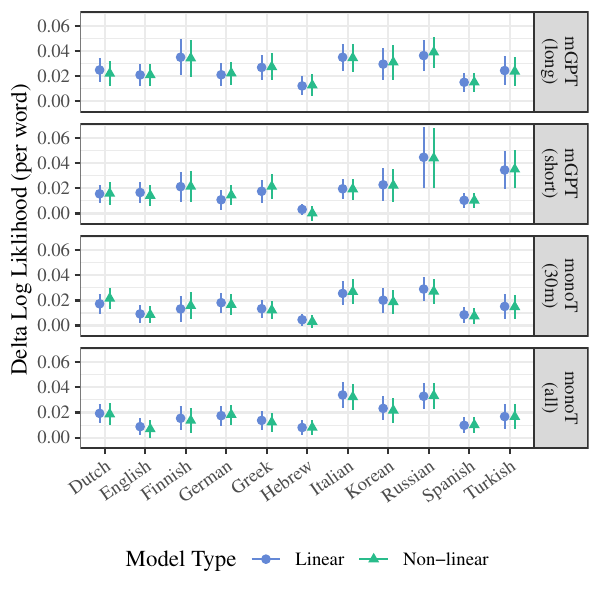}
    \end{minipage}
    \vspace{-0.5cm}
    \caption{\textbf{Comparison Between Linear and Non-linear Models:} Error bars are 95\% CIs of \dll. Results are for gaze duration. We observe no difference between non-linear GAMs (green) and linear GAMs (blue) across languages.}
    \label{fig:linearity-comp}
    \vspace{-10pt}
\end{figure}

The results of this comparison can be seen in Figure \ref{fig:linearity-comp}. Here, \dll is slightly different for linear models than in Section \ref{sec:surp-result}, as we fit these models with tensor product terms for baseline predictors. Visually, there is no consistent difference between linear and non-linear models across languages. 
We test the difference in \dll statistically with permutation tests, as described in Section \ref{sec:regression_models}.
Our tests do not support the alternative hypothesis for an $\alpha=0.05$ for any of the models or languages. Together with the visualizations presented above, these results support a linear linking function between surprisal and reading times.\looseness=-1

\section{Discussion}

\subsection{Implications of Psycholinguistic Theories}

Throughout the paper, we have mentioned that the eleven languages studied come from five different language families, but what does this mean in terms of the actual linguistic characteristics that they exhibit? At the highest organizational level, our sample includes languages with multiple different word orders and headedness including SVO (Hebrew, English), SOV (Korean, Turkish), as well as languages with no dominant word order \citep[German and Greek;][]{haspelmath2005world}. Our sample includes languages with extensive case marking such as Finnish (15 cases), as well as languages with extremely impoverished case systems, such as English. In terms of word construction, our sample includes languages that are both agglutinating (Turkish, Finnish and Korean) and fusional (Russian, Romance languages). While this set is not close to covering all ways that human languages can vary, we bring up these differences to highlight how it does contain important high-level parametric variations observed in human languages.

In light of this, the stability observed in our results testing the surprisal hypothesis is rather remarkable. Across language families and model types, we observe essentially consistent results, in terms of the predictive power of the models, the effect size associated with surprisal, as well as for the shape of the surprisal--reading-time relationship. Focusing first on predictive power, we find a relatively tight range of \dll values associated with surprisal. For example, for gaze duration and mGPT, all \dll values fall between $0.012$ and $0.040$. Indeed, across languages and models, we find relatively little variance in the predictive power of surprisal. Turning to the effect size of surprisal, we observe a millisecond-per-bit trade-off that falls between 2--4 ms/bit for every language (See Figure \ref{fig:coeffs}). The previous estimate of $3.75$ ms of slowdown per bit of surprisal reported in \citet{smith2013effect} for English falls well within this range (though note that this previous work used surprisal estimates derived from an $n$-gram model, which will generally be higher than surprisal estimates derived from large neural language models such as the ones we consider in this study).
We take these results to suggest that humans may have stable crosslinguistic preferences for the rate at which they process information during reading, i.e., not greater than 4 milliseconds per bit of information.
This is consistent with previous work that has observed crosslinguistic consistency in the rate of information during speech production \citep{pellegrino2011cross, coupe2019different}, as well as trade-offs between the information content of a word and the time taken to produce it \citep{pimentel-etal-2021-surprisal}.\footnote{Our results are not necessarily consistent with a universal channel \emph{capacity}, or an information rate above which comprehension cannot be sustained. A channel capacity could explain uniform information density effects, or the tendency to spread information out uniformly over a sentence, presumably at or near the channel capacity \citep{jaeger2006speakers, frank2008speaking, meister2021revisiting}. However, as pointed out in \citet{smith2013effect}, such effects require a superlinear surprisal link hypothesis, which we do not observe empirically.\looseness=-1}\looseness=-1

One point of difference between these and previous results, however, is the size of the effect of the surprisal of previous words. Looking at gaze duration in the Dundee corpus of English \citep{kennedy2003dundee}, \citet{smith2013effect} find an effect on reading time for surprisal for the previous word which is about as strong as for the current word. We find much weaker effects in this study, ranging from 0-2 ms/bit. Note, that this lower effect for previous words is in line with other incremental processing measures which are strongly incremental, such as the maze task, where previous-word surprisal has little to no effect on reading time of the current word \citep{boyce2020amaze}, as well as with the results reported in \citet{pimentel2022effect} for eye-tracking over the Provo \citep{luke2018provo} and Dundee corpora.

Turning to the shape of the surprisal--reading times relationship, our results support the linear link hypothesis and are in line with the comprehensive results recently reported in \citet{shain2022large}. Unlike \citet{hoover2022plausibility} we do not observe superlinear surprisal--reading time relationships for larger and more data-intensive language models, or for language models that had access to longer contextual windows. Interestingly, we do observe that the one language which visually appears to be superlinear, i.e., it has an upwards curve in Figure \ref{fig:gam-surp}) is English. 
Thus, while we believe \citet{hoover2022plausibility} was right to be concerned by a potential visual nonlinearity in the English relationship, this effect does not appear to exist crosslinguistically and is not borne out by our statistical testing.\looseness=-1

Surprisal theory is attractive because it offers a general-purpose link between statistical properties of natural language and human behavior. 
While its domain generality gives the theory a universal-like flavor, previous literature has (in our opinion) correctly refrained from overtly discussing it as a universal of human language processing. 
By conducing the most comprehensive crosslinguistic assessment of surprisal theory to date, this study presents initial evidence which supports the universality of surprisal effects in naturalistic reading.
That being said, further testing is a necessary next step.\looseness=-1

\subsection{Implications of Multilinguality}

As the number of multilingual language models has proliferated, it has become increasingly important to understand how they differ from more traditional, monolingual models. Previous studies have produced mixed results: Some have found that the larger training data scales of multilingual models leads to better performance \citep{conneau-etal-2020-unsupervised}, while others have found advantages for monolingual models \citep{agerri-etal-2020-give, ronnqvist-etal-2019-multilingual, Virtanen2019MultilingualIN}, which are often attributed to monolingual model's language-specific tokenization and vocabulary representation.
The majority of these previous studies have focused on masked language models (mostly using architecture based off the BERT model) and evaluation based on performance of downstream tasks \citep{doddapaneni2021primer}. This study offers a useful complement to previous work by focusing  on autoregressive models, as well as on their cognitive modeling capacities.\footnote{However, see \citet{hollenstein-etal-2021-multilingual} for a previous investigation of multilingual language models' ability to predict reading times.\looseness=-1}
Our results are more or less in line with previous studies, insofar as we find no obvious differences between our multilingual model and our monolingual models. Our results thus suggest that for computational linguists interested in cognitive modeling, multilingual and monolingual language models may be equally viable options. However, we would like to note that we did not compare models in truly low-resource settings, as the  training datasets of our smallest monolingual models still included 30 million tokens. It may be the case that when trained on much smaller datasets, multilingual models may benefit from crosslingual transfer.

\subsection{Concurrent Work} 

We want to briefly note the differences between the work presented here and a concurrent study that also used the MECO  dataset \citep[i.e.,][]{de2022effects}. While \citeauthor{de2022effects}'s research questions are similar to ours, their methods and conclusions are quite different. Instead of an autoregressive language model, they use a masked language model \citep[mBERT;][]{devlin2018bert}, which has access to both left and right context. 
An issue with this strategy is that the surprisal values produced by this setup are not psychologically plausible estimates of actual surprisals, which are estimated from the left context alone,\footnote{Because the perceptual span is limited to about 14 characters to the right of a fixation \citep{rayner:1975} and little linguistic information is gleaned from the far right of the perceptual span \citep{schotter-etal:2012-parafoveal}, upcoming word identities cannot have a substantial causal influence on a word's first-pass reading behavior \citep{granger1969investigating}.} which weakens the ability to test psycholinguistic causal claim about the relationship between surprisal and reading times. In their experiments, \citeauthor{de2022effects} do not find significant effects of pseudo-surprisal on gaze duration in four of the 12 languages in MECO,\footnote{They include Estonian, which we drop as it was not in mGPT's training data.} including English, and find significant effects of pseudo-surprisal on other eye movement measures in even fewer of the languages, which they view as evidence that surprisal might \textit{not} be a consistent predictor of reading times across languages.\footnote{Their study does not consider contextual entropy.} 
While we are aligned on the importance of \citeauthor{de2022effects}'s research questions, we believe that their failure to replicate surprisal effects for English---or to find it for other languages---reflects the limitations in their methodological choices.\looseness=-1

\subsection{Limitations and Future Directions} 

Turning back to our own study, there are a few limitations we would like to discuss: Although our sample of languages is much larger than previous studies, Indo-European languages are still overrepresented. 
Indeed, each of our non Indo-European language families is represented by a single language. Additionally, all the data tested here comes from high-resource languages with long traditions of writing systems, and from individuals who live in industrialized societies. Finally, the methodology we employ here requires a large corpus of (written) language on which a language model can be trained. It may be the case, that for much lower-resource languages, there is often not enough linguistic data to derive statistical estimates needed to test surprisal theory in this manner.
Thus, while our methods may be able to test the predictions of surprisal theory in lower-resource settings, where corpora of a few hundred thousand words exist, they may not be suitable for a large number of the world's languages. 
While our results put surprisal theory on firmer empirical footing, testing its predictions beyond these settings is an important and necessary step in assessing the theory's universality.\looseness=-1

\section{Conclusion}

This paper has presented the most comprehensive crosslinguistic evaluation of surprisal theory reported in the literature to date.
Using eye-tracking data from controlled materials in eleven languages across five language families, we have tested three hypotheses: (i) the surprisal hypothesis (surprisal is predictive of reading times), (ii) the contextual entropy hypothesis (contextual entropy is predictive of reading times), and (iii) the linear link hypothesis (the relationship between surprisal and reading times is linear). We found exceptionally strong crosslinguistic stability in our results, with each prediction being borne out in every language tested. These results provide the most robust link between information-theoretic quantities and incremental processing.\looseness=-1

\section*{Acknowledgments}

We would like to thank our TACL action editor, Maggie Li, as well as our reviewers, whose thoughtful feedback greatly improved this work. 
Tiago was supported by a Facebook PhD Fellowship.
Clara was supported by the Google PhD Fellowship.
Ethan was supported by an ETH Zurich Postdoctoral Fellowship. 
Roger was supported by NSF grant BCS-2121074 and a Newton Brain Science Award.\looseness=-1


\bibliography{everything}
\bibliographystyle{acl_natbib}

\onecolumn
\appendix

\section{Regression Modeling Details} \label{app:regressions}

\lstset{frame=none,
  language=R,
  showstringspaces=false,
  columns=flexible,
  numbers=none,
  keywords={lmer, data, te, bs, s, gam},
  basicstyle={\small\ttfamily},
  keywordstyle=\color{Blue},
  stringstyle=\color{DarkGreen},
  commentstyle=\color{DarkGreen},
  breaklines=true,
  breakatwhitespace=true,
  tabsize=3}

\newcommand{\listlingr}[1]{{\small\ttfamily#1}}
We give more details on the regression formulae used in the various experiments reported in the main section. 
Our notation is as follows: \listlingr{reading\_time} is the reading time of the word of interest, i.e., $w_t$, \listlingr{surp} is the surprisal of $w_t$, \listlingr{prev\_surp} is the surprisal of the previous word, i.e., $w_{t-1}$, and \listlingr{prev2\_surp} is the surprisal of the word two previous, i.e., $w_{t-2}$. 
The other variables use the same \listlingr{prev} and \listlingr{prev2} prefixes, and we simply explain the variable names for the current index $t$ for the sake of brevity, below. 
For these, \listlingr{ent} indicates the contextual entropy of $W_t$, \listlingr{len} indicates the length of $w_t$ in characters, and \listlingr{freq} indicates the negative log 
 frequency of $w_t$.

\paragraph{Effect of Surprisal (Section \ref{sec:surp-result})} 
For the tests assessing the effect of surprisal within individual languages, we use the following model:
\begin{lstlisting}
lmer(reading_time ~ surp + prev_surp + prev2_surp + freq + len + prev_freq + prev_len + prev2_freq + prev2_len, data = .)
\end{lstlisting}
The baseline models are the same with the exception that the surprisal terms are removed. 
For the aggregate test assessing the effect of surprisal across languages, we use the following model:
\begin{lstlisting}
lmer(reading_time ~ surp + prev_surp + prev2_surp + freq + len + prev_freq + prev_len + prev2_freq + prev2_len + (surp + prev_surp + prev2_surp + freq + len + prev_freq + prev_len + prev2_freq + prev2_len | lang), data = .)
\end{lstlisting}

\paragraph{Effect of Contextual Entropy (Section \ref{sec:entropy})} 
For both tests, the baseline model included surprisal, length and negative log frequency, i.e., it was the first model given in the paragraph above. 
For the \emph{replace} test, the target regression model we use is
\begin{lstlisting}
lmer(reading_time ~ ent + prev_ent + prev2_ent + freq + len + prev_freq + prev_len + prev2_freq + prev2_len, data = .)
\end{lstlisting}
For the \emph{add} test, the target regression model we use is
\begin{lstlisting}
lmer(reading_time ~ ent + prev_ent + prev2_ent + surp + prev_surp + prev2_surp + freq + len + prev_freq + prev_len + prev2_freq + prev2_len, data = .)
\end{lstlisting}

\paragraph{Surprisal--RT Linking Function (Section \ref{sec:rt-surp-link})}
The GAM formula used for non-linear models we use is
\begin{lstlisting}
gam(reading_time ~ s(surp, bs = 'cr', k = 6) + s(prev_surp, bs = 'cr', k = 6) + te(freq, len, bs = 'cr') + te(prev_freq, prev_len, bs = 'cr'), data = .)
\end{lstlisting}
And for linear models:
\begin{lstlisting}
gam(reading_time ~ surp + prev_surp + te(freq, len, bs = 'cr') + te(prev_freq, prev_len, bs = 'cr'), data = .)
\end{lstlisting}
We now briefly explain the components of these regressions. \listlingr{s()} sets up a spline-based smooth term between a predictor and response variable that can take on a wide variety of non-linear functional relationships. 
Here, \listlingr{k=6} indicates a maximum of 6 basis functions for the smooth.
We choose \listlingr{k=6} following the logic from \citet{hoover2022plausibility}, Appendix C. 
Having 6 basis functions allows for five degrees of freedom, which enables the regression to fit non-linear yet still relatively simple curves. 
The other term, \listlingr{te()}, sets up a tensor product smooth term, which can effectively capture non-linear interactions between two variables.

\begin{figure*}
    \centering
    \begin{minipage}{\textwidth}
    \centering
    \includegraphics[width=\textwidth]{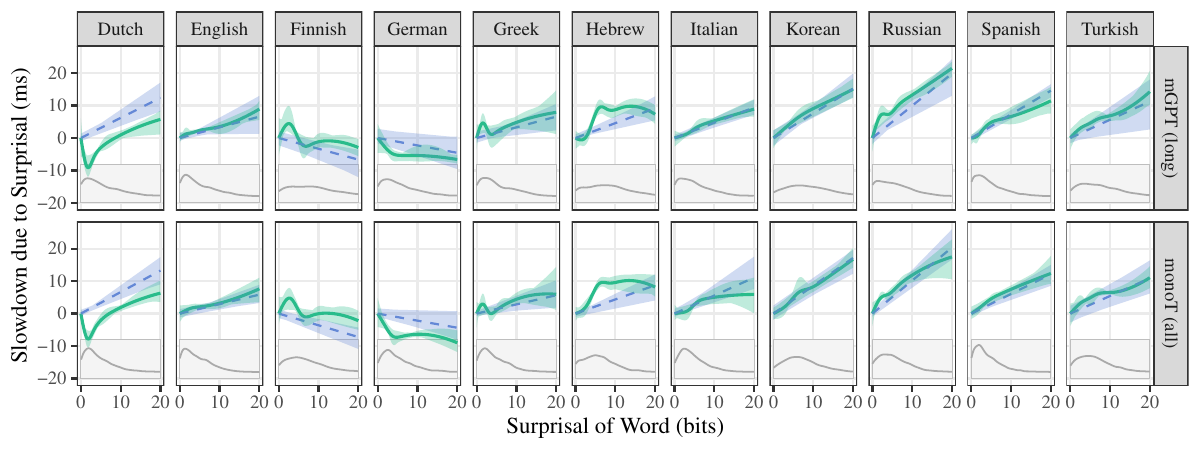}    \end{minipage}
    \caption{\textbf{Surprisal versus Reading Time Relationship (Previous Word):} Non-linear GAMs are in green and linear control GAMs are in dotted blue. Results are for gaze duration. Shaded regions represent bootstrapped 95\% confidence intervals. Grey subplots indicate the distribution of surprisal values.}
    \label{fig:gam-surp-prev}
\end{figure*}

\section{Surprisal versus RT for ${w_{t-1}}$ }

As mentioned in the main text, previous work has investigated the relationship between surprisal and reading times not just for the current word $w_t$, but also for the previous word, $w_{t-1}$. 
Looking at gaze duration in the Dundee corpus of English \citep{kennedy2003dundee}, \citet{smith2013effect} find an effect of $w_{t-1}$'s surprisal which is about as strong as the effect of $w_t$'s surprisal on the reading time of $w_t$. 
In Figure \ref{fig:gam-surp-prev} we show this relationship in our corpus for mGPT and monoT(all), using the same methods and presentational paradigm as in Section \ref{sec:gams}. 

The results are consistent across models used, and suggest that the relationship between reading time and surprisal of the previous word is somewhat variable across languages. For English, Italian, Korean, Russian, Greek and Spanish we find a relationship that is roughly linear and increasing, i.e., similar to the results for surprisal of the current word. For Dutch and Turkish, we find a relationship that is roughly increasing, but visually non-linear. For Finnish, German and Hebrew, we find either a negative relationship, or a relationship that is difficult to interpret. These results are in line with the effect terms plotted in Figure \ref{fig:coeffs}, where we find very weak and sometimes negative coefficients for the $w_{t-1}$ surprisal term for these languages (i.e., the middle $x$-tick position in the top row). Overall, these results are consistent with the linear effect that has been previously observed in English. However, they suggest that the impact of the surprisal of the previous word varies between languages.

\end{document}